# GraphFire-X: Physics-Informed Graph Attention Networks and Structural Gradient Boosting for Building-Scale Wildfire Preparedness at the Wildland-Urban Interface

*Miguel Esparza[1*], Vamshi Battal[2], and Ali Mostafavi[1]*

[1] Urban Reslience.AI Lab, Zachry Department of Civil and Environmental Engineering, Texas A&M University, College Station, TX 77840, United States of America

[2] Department of Computer Science and Engineering, Texas A&M University, College Station, TX 77840, United States of America

*Corresponding Author:* mte1224@tamu.edu

## Abstract

As wildfires increasingly evolve into urban conflagrations, traditional risk models that treat structures as isolated assets fail to capture the non-linear contagion dynamics characteristic of the wildland–urban interface (WUI). This research bridges the gap between mechanistic physics and data-driven learning by establishing a novel dual-specialist ensemble framework that disentangles vulnerability into two distinct vectors: environmental contagion and structural fragility. The architecture integrates two specialized predictive streams: an environmental specialist, implemented as a graph neural network (GNN) that operationalizes the community as a directed contagion graph weighted by physics-informed convection, radiation, and ember probabilities, and enriched with high-dimensional Google AlphaEarth Foundation embeddings; and a Structural Specialist, implemented via XGBoost to isolate granular asset-level resilience. Applied to the 2025 Eaton Fire, the framework reveals a critical dichotomy in risk drivers. The GNN demonstrates that neighborhood-scale environmental pressure overwhelmingly dominates intrinsic structural features in defining propagation pathways, while the XGBoost model identifies eaves as the primary micro-scale ingress vector. By synthesizing these divergent signals through logistic stacking, the ensemble achieves robust classification and generates a diagnostic risk topology. This capability empowers decision-makers to move beyond binary loss prediction and precisely target mitigation—prioritizing vegetation management for high-connectivity clusters and structural hardening for architecturally vulnerable nodes—thereby operationalizing a proactive, data-driven approach to community resilience.

**Keywords:** *Wildland–Urban Interface (WUI); Physics-informed Machine Learning; Graph Neural Networks; Contagion Dynamics; Precision Mitigation; AlphaEarth Foundation Embeddings.*

## 1.0 Introduction

The trajectory of recent wildfire season is alarming, evolving from natural ecological events into uncontrollable urban conflagrations that devastate the built environment. The consequences of these wildfires, especially when interacting with communities, have been dire and are the culprit of socio-economic losses all around the world. Particularly, events such as the 2023 Lahaina disaster - the deadliest in Hawaii's history, claiming 102 lives and inflicting $5.5 billion in damages (Insurance Institute for Business and Home Saftey, 2024). This threat was further exemplified by the January 2025 Eaton and Palisades fires in Los Angeles County, which destroyed approximately 16,000 structures. The economic impact of these two events was estimated at $40 billion, representing 80% of the average annual global cost of wildfires in 2023 (Mahmoud, 2024). Climate change increases the frequency of conflagrations, yet communities continue to develop near forested areas (Mahmoud, 2024). As urbanization expands deeper into fire-prone regions, the scale of destruction is increasing globally (Chulahwat et al., 2022; Mahmoud & Chulahwat, 2020; Taccaliti et al., 2023). Despite mitigation strategies,. The demand to mitigate vulnerabilities has uncovered an urgent need to assess wildfire risk at the building level; however, limited work exists on predicting fire risk for individual structures using data-driven (Madaio et al., 2016). Prior research assessing urban fire risk has allocated buildings into municipal blocks rather than examining them individually (Clare et al., 2012; DaCosta et al., 2015). Building-level risk assessment can enhance mitigation strategies by enabling decision-makers to develop tailored plans that ensure community resilience. These plans can be further enhanced through machine learning and deep learning methods capable of capturing complex environmental dynamics and identifying patterns among individual buildings.

Despite rapid advances in wildfire simulation and data-driven prediction, a critical gap persists at the structure scale where preparedness and protective action decisions are made. Physics-based spread models are often computationally intensive and oriented toward wildland behavior, while many machine learning and deep learning approaches prioritize regional fire occurrence or next-day spread rather than building-level damage. Moreover, these data-driven methods frequently treat structures as independent observations, overlooking the inter-building and vegetation contagion pathways that drive Wildland-Urban Interface (WUI) losses. Existing building-focused studies either rely primarily on tabular structural attributes (e.g., roofing, eaves, vents) without representing the surrounding environmental exposure at comparable fidelity, or employ graph representations but limit themselves to traditional network metrics, constraining predictive power and failing to leverage high-dimensional geospatial signals. This "spatial blindness" of current approaches ignores the physical reality that WUI fires are contagion events where a building's survival is inextricably linked to its neighbors and surrounding topography.

Theoretically, this research challenges the prevailing atomistic paradigm of wildfire risk assessment, which mischaracterizes structures as independent assets governed solely by their material hardening. We posit that urban conflagrations are not static exposure events but dynamic network diffusion processes in which the built environment itself serves as the primary transmission vector. By formalizing the WUI as a directed graph weighted by mechanistic pathways of convection, radiation, and ember transport, this framework captures the non-linear cascade dynamics essential for explaining "cluster failures" in high-density areas. This conceptual shift—from analyzing intrinsic susceptibility (a building's design) to topological vulnerability (a building's connectivity)—bridges a critical gap in resilience theory, providing a rigorous mathematical basis for understanding how structurally robust buildings can succumb not to direct wildland impact, but to cumulative "thermal pressure" propagated through the failure of neighboring structures.

From a practical perspective, fire risk management practitioners are hindered by a resolution gap in which mitigation and preparedness strategies rely on coarse, aggregate hazard mapping rather than granular, building-specific diagnostics. Current decision-support tools often function as black boxes, designating areas as high-risk without distinguishing whether vulnerability stems from broad landscape exposure (requiring vegetation management) or specific architectural deficiencies (requiring structural hardening). This lack of diagnostic clarity leads to mitigation paralysis, where decision-makers struggle to prioritize limited resources effectively.

This research addresses these limitations by developing a Dual-Specialist ensemble framework (called GraphFire-X) that unifies competing scales of analysis. The rationale for this approach rests on the understanding that WUI fire vulnerability is driven by two distinct but interacting vectors: environmental contagion (the probability of fire propagating to a location via convection, radiation, and embers) and structural fragility (the probability of a building igniting once exposed). The framework integrates a physics-informed Graph Neural Network (GNN) that captures non-linear, neighborhood-level spread dynamics with a gradient-boosted decision tree (XGBoost) optimized for granular building features. The GNN incorporates high-dimensional environmental features that would be difficult for tabular models to process, while the XGBoost model excels at processing structural attributes in tabular format. The damage probability outputs from both models serve as inputs for logistic regression stacking, enabling the framework to disentangle environmental risk from structural fragility and support targeted, pre-disaster interventions. The proposed methodology is applied to the Eaton Fire in California. This fire ignited on January 7, 2025, in the Altadena/Pasadena area of Los Angeles County, destroying more than 7,000 structures, burning more than 14,000 acres, and claiming 16 lives. The fire intensified due to strong winds, extreme drought conditions, and tight spacing between buildings.

This research advances wildfire risk management by establishing a novel ensemble architecture that bridges the methodological gap between deterministic physical simulations and stochastic deep learning. By embedding mechanistic heat transfer equations—governing convection, radiation, and ember spotting—directly into the topology of a GNN and enriching it with high-dimensional Google AlphaEarth Foundation embeddings, the model captures the complex, non-linear spatial dependencies of fire spread that tabular methods often miss. The study demonstrates that robust vulnerability assessment requires the simultaneous integration of broad, landscape-level environmental contexts with granular, building-level structural features, offering a scalable blueprint for hybridizing domain physics with advanced representation learning. This framework shifts the paradigm from binary loss prediction to actionable, cause-specific mitigation planning. Neighborhoods flagged by the GNN can be prioritized for community-scale vegetation management and firebreak construction, while properties identified by the XGBoost model can be targeted for structural hardening incentives, such as retrofitting eaves and vents. This granular vulnerability triage empowers stakeholders—from urban planners to insurers—to implement precise, data-driven resilience strategies before ignition occurs.

The remainder of this paper is organized as follows. Section 2 reviews current studies assessing wildfire behavior, recent machine learning and deep learning frameworks, and graph network applications at the building level. Section 3 describes the data used, encompassing high-dimensional satellite embeddings, building features, and vegetation characteristics. Section 4 details the methodology, including the adopted graph theory from Chulahwat et al. (2022), the GNN architecture, the XGBoost model, and the probabilistic fusion via logistic regression stacking. Section 5 presents results and discussion, followed by a summary of findings, limitations, and avenues for future research in Section 6.

## 2.0 Literature review

The challenge of predicting wildfire behavior, intensified by climate change, has prompted the development of various modeling approaches. These approaches range from simulated to analytical (Sullivan, 2009). For example, simulated models such as BehavePlus (Andrews, 2006), FARSITE (Mark Arnold Finney, 1998), FSPro(Mark A Finney et al., 2011), and national Bushfire model (Knight & Coleman, 1993) examine fire behavior and growth under various dynamic scenarios related to weather and topography. Analytical methods like FlameMap (Mark A Finney, 2006; Stratton, 2006) and Rothermel's model (Rothermel, 1972) provide spatial mapping of fire spread for a single point or entire landscape. While these models are widely accepted and cover many aspects of wildfires, they are computationally expensive and tend to focus exclusively on wildlands or on localized fire propagation within communities, making them less generable(Jiang et al., 2022). However, recent advances in the availability of Earth observation data (Brown et al., 2025; Zhao et al., 2024) , artificial intelligence methods (Dargin et al., 2020; Esparza et al., 2025; Luo & Lian, 2024; Xiao et al., 2025), and computation resources(Esparza et al., 2025; Wu et al., 2020) have paved the way for novel models that have the potential to overcome the accuracy and transferability issues of conventional wildfire behavior methods, thus making them more generalizable (Cardil et al., 2021; Jain et al., 2020; Rösch et al., 2024).

ML and DL methods have allowed researchers to explore data-driven approaches that can enhance speed and efficiency when predicting wildfire behavior. For example, support vector machines (SVM) were used by Rahman et al. (2022) for early fire detection system and emergency management. The researchers used large datasets of fire videos from the internet which were able to achieve an accuracy of 93.33% when predicting real-time fire detection (Rahman et al., 2022). Tehrany et al. (2021) created a logistic regression model to predict the risk of wildfire in the Brisbane Catchment of Australia and achieved an overall accuracy of 90% (Tehrany et al., 2021). To capture more complex patterns and relationships in the wildfire domain, researchers have used deep learning (Boroujeni et al., 2024; Dabrowski et al., 2023; Xu et al., 2024). For example, Qiao et al. (2024) used a transformer-based neural network to identify the ignition point by simulating the backward spread of wildfire spread. (Qiao et al., 2024). Hou et al. (2022) developed a convolutional neural network based on environmental variables to predict wildfire spread in small test regions (Huot et al., 2022). Kadir et al. (2023) used historical wildfire data for an long short-term memory (LSTM) model to predict the yearly spatiotemporal distributions of wildfires in Indonesia and achieved a success rate better than 90% (Kadir et al., 2023). Cao et al. (2024) employed a transformer to predict next-day wildfire risk in Quanzhou County, Guangxi Province, China, based on three days of meteorological, topographical, and human activity data. The model's performance, outperformed LSTM, recurrent neural network (RNN) and support vector machine (SVM) in terms of generalization (Cao et al., 2024). Machine learning (ML) and deep learning (DL) offer promising potential in terms of speed and transferability; however, they face challenges related to data quality and model interpretability. To this end, researchers have explored other avenues to model wildfire behavior, particularly with graph theory. Graphs consist of a series of nodes and edges that can characterize complex interactions in the physical world, such as in road networks (Dong et al., 2022; Esparza et al., 2021, 2024), flood risk (Dong et al., 2020; Farahmand et al., 2022, 2023), and disease spread (Alguliyev et al., 2021; Ma et al., 2025; Shirley & Rushton, 2005; Tomy et al., 2022). The relationships between graph nodes have the potential to characterize wildfire spread, enhancing insights from data-driven approaches and improving interpretability. Graph theory has been combined with neural networks to form graph neural networks (GNNs). Jiang et al. (2022) further advanced graph-based modeling by introducing an irregular graph network (IGN), using dense nodes for complex regions and sparse nodes for simpler ones(Jiang et al., 2022). This variable-scale approach ensures that graph edges maintain homogeneous properties regarding fuel and topography. Wildfire spread is then simulated using a deep neural network that calculates spread time, flame length, and fire intensity along graph edges. This method explicitly demonstrates the spatiotemporal spread route, offering a novel way to visualize and analyze fire dynamics. Rosch et al. (2022) used a spatial graph neural network, trained on a time-series dataset of wildfires from 2016 to 2022, to predict the daily spread of European wildfires. The models for Portugal and the entire Mediterranean achieved a 0.37 and 0.36 intersection over union metric, which highlights the complexity of next-day fire prediction(Rösch et al., 2024).

To evaluate households during a natural hazard, Noumeur (2025) used random forest and XGBoost models to classify structural damage classification caused by wildfires using building features (Noumer, 2025). Both models achieved 90% accuracy and found that feature-important analysis reaffirms that newer, higher-value homes and fire-resistant construction features, such as enclosed eaves and window vents, substantially reduce wildfire damage risk. Tree-based models were used to classify tornado building damage. Danrong Zhang et al. (2025) used a decision tree, random forest, XGBoost, and a gradient-boosting model to classify buildings into five categories ranging from no damage to destroyed (Zhang et al., 2025). Of these three, the most accurate model was the random forest, with an accuracy of 52%, a precision score of 53.32%, and F1 score of 48.66%. The random forest model identified year of construction of the building and the tornado's path as important features for damage assessment. This study also examined GNN methods to predict building damage to account for debris from a damaged building. The GNN method enables assessment of the interactions of buildings during a tornado rather than treating them as single entity. This type of internation is important as it produced an 83.53% accuracy.

Modeling the buildings as a network rather than as a single entity can capture the mechanism of fire spreading from building to building, thus providing more robust insights. Mahmoud & Chulahwat (2018) developed a framework that accounts for community-specific layouts, wind conditions, structural features, and vegetation to quantify vulnerability of structures (Mahmoud & Chulahwat, 2018). The edges are a component of fire mechanisms, such as convection radiation and ember spotting. This framework found that building risk depends on meteorological conditions, environmental factors, community characteristics, and layout. To expand on the framework, Mahmoud & Chulahwat (2018) developed two metrics, a modified degree formulation that assesses a node's vulnerability as the mean of incoming edge weights and the modified random walk formulation. In the first metric, the edge weights represent the probability of a fire spreading from node to node. The second metric evaluates node vulnerability based on the fire transmissibility of a neighboring node. They applied these metrics to the 2018 Camp Fire and 2020 Glass Fire and achieved prediction accuracy ranging from 58% to 64%. The study concluded that a building's survival is strongly correlated with the fire behavior of its immediate neighbors (a one-degree separation), underscoring the importance of local interactions (Chulahwat et al., 2022).

## 3.0 Datasets

To train both the GNN and XGB (XGBoost) models, a binary structural damage outcome of the wildfire is incorporated as the ground truth (Figure 1). Figure 1a shows the predictive features for the GNN, a high-dimensional feature set related to vegetation, climate, topography, and weather, as well as to structural features of buildings. The associated features for each node and dataset can be further examined in Appendix B. The topographic slope and elevation for the buildings and vegetation are also considered as features for the GNN. Other inputs to GNN necessary to support the graph computation are vegetation nodes, fuel model, and area and volume (the calculation is explained in Appendix B) of buildings. Data used by the XGB model are vegetation data, slope, elevation, fuel model, area and volume, and structural features (Figure 1b).

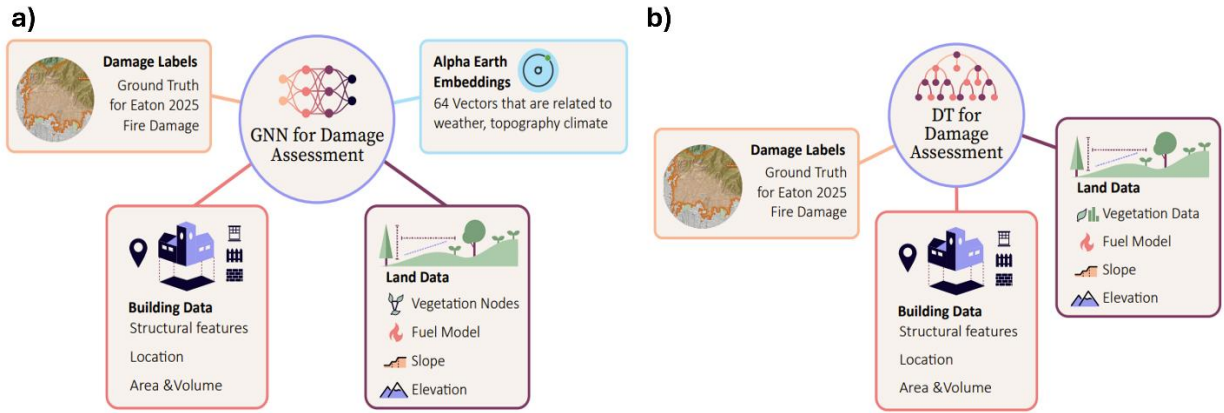

**Figure 1:** *Datasets for the (a) Graph neural network model and (B) XGBoost model. The GNN model has damage labels, Google AlphaEarth Foundation embeddings, structural features, slope and elevation for both the building nodes and vegetation nodes. The XGB model has damage labels, structural features, vegetation data, slope, elevation, and fuel model.*

The data used in this study, detailed in Sections 3.1 through 3.3, integrates diverse measurement modalities necessary for robust wildfire vulnerability assessment. Section 3.1 highlights the California Natural Resource Agency Damage Inspection database (DINS). This dataset incorporates the ground truth label from the Eaton fire and structural features that are used as predictors. Microsoft building footprint data was subsequently incorporated to obtain building area and volume. Section 3.2 introduces the novel, high-dimensional Google AlphaEarth Foundation (GAEF) dataset, implemented exclusively within the GNN architecture due to its capacity to process dense feature sets. This dataset incorporates 64 vector embeddings encompassing long-term climate, weather, topography, and vegetation features. Critically, the GAEF data aggregates information across the entirety of 2024, capturing the essential months-long buildup of drought and heat conditions relevant to wildfire ignition and spread. Due to this high dimensionality, the GAEF dataset is omitted from the XGB model. To capture vegetation characteristics in this model, 2024 LandFire data was used as outlined in section 3.3

## 3.1 Building dataset

The Eaton fire damage data comes from the California Natural Resource Agency Damage Inspection database with roughly 18,000 structures classified into five categories: (1) no damage (2) affected, (3) minor, (4), major, and (5) destroyed. This research uses a binary classification of structure damage; therefore, all four damage categories are combined into one class. Combining the DINS data into binary variables has been used in past studies that implement building-level damage assessment for graph problems (Chulahwat et al., 2022). The rationale is that during the preparedness phase of a wildfire, identifying potentially vulnerable structures is critical. During this phase, misclassifying a partially damaged structure as destroyed or slightly damaged, is less critical than misclassifying a non-damaged building or completely destroyed property entirely (Noumer, 2025). There are studies that have incorporated multiple damage categories when estimating building damage from natural hazards such as wildfire or tornado(Esparza, Gupta, et al., 2025; Luo & Lian, 2024). These studies indicate that the intermediate damage categories provide a degree of imbalance which leads to their ML and DL models to have lower classification metrics compared to the no damage and destroyed categories(Esparza, Gupta, et al., 2025; Luo & Lian, 2024; Noumer, 2025; Zhang et al., 2025). There are other methods that can improve granular damage classification which typically reside in the post-event phase. Methods such as vision language models (Esparza, Gupta, et al., 2025; Xiao et al., 2025) and vision image transformers (Luo & Lian, 2024) have proven to be suitable methods for post-damage classification where it is more critical to have granular damage estimation for resource allocation and insurance estimation. However, since this study focuses on the preparedness phase of a wildfire event, this research used binary classification, for both the GNN and XGB, since either damage outcome will require some level of response, whether to repair or completely rebuild the structure.

In addition to the damage labels, the DINS database contains structural features that become inputs for both the GNN and XGB model's predictions on the binary damage classification task. These structural features, if applicable to the building, include the roofing, siding, deck, porch and fence materials, as well as the presence of eaves, windows, and vent screening. These features are encoded based on their material type and presence on the structure to be functional inputs in the model. To include the building area and volume, Microsoft building footprint data was allocated to the structures. The volume and area will be used as calculation inputs for the GNN's edge weights, and act as feature inputs for the XGB model.

## 3.2 Google AlphaEarth Foundation

The GAEF dataset integrates environmental risk attributes into a universal feature space. The GAEF employs a 64-dimensional vector embedding field that compactly summarizes environmental and climate conditions for every 10-meter plot of land. These vectors are assigned to the closest building node when constructing the GNN. The vector embeddings encompass are summarized below:

*Topographic features:* The GLO-30 and GEDI sensors were used to gather elevation, slope and canopy structure.

*Climate and weather features:* Drought conditions, seasonal temperature curves, precipitation data were obtained from ERA5-LAND.

*Fuel Characteristics features:* Vegetation type, plant health, and moisture content are from optical and radar sensors.

The model was trained using a self-supervised teacher-student framework that learns to reconstruct masked multi-sensor data into consistent embeddings while aligning them with geotagged text to capture semantic meaning. To ensure robustness, the system minimizes the difference between a 'teacher' model viewing full inputs and a 'student' model viewing sparse or corrupted data, forcing the network to learn stable representations of the land surface despite missing observations. The resulting vectors were validated by feeding frozen embeddings into simple linear and k-nearest neighbor classifiers across diverse downstream tasks, such as crop mapping and biophysical variable estimation, where they consistently outperform existing bespoke models (Brown et al., 2025).

The GAEF dataset is purely used on the GNN due to its ability to process high-dimensional data unlike the XGB model. This will treat the GNN as an environmental specialist, since the focus of this model is the 64-vector embeddings, while the XGB model will act as a structural specialist due to its ability to process tabular features effectively.

### 3.3 LandFire dataset

The LandFire database supplies the fuel and vegetation data, gathered from LiDAR and satellite inputs. The data include canopy bulk density (CBD), canopy base height (CBH), canopy height (CH), and canopy density (CD), which provide insight into the vegetation landscape. Additionally, LandFire utilizes the standardized 40 Scott and Burgan fire behavior fuel model (FBFM40). A fuel model is not raw data, but rather a standardized set of parameters that quantitatively describe a fuel bed, specifying load, depth, heat content, and moisture of extinction. For the GNN model, the FBFM40 data are used to determine the necessary fire characteristics to implement the graph methodology by Chulahwat et al. (2022) For the XGB model to have vegetation data to complement the structural features, CBD, CBH, CH, CD, Fuel Model are incorporated into the model.

## 4.0 Methodology

This methodological framework for robust property-level wildfire damage prediction focuses on two models, a GNN and XGB. The damage probabilities for buildings in the study area are input into a logistic regression model to examine classification results when considering both the GNN and XGB models. The proposed methodological workflow is summarized in Figure 2.

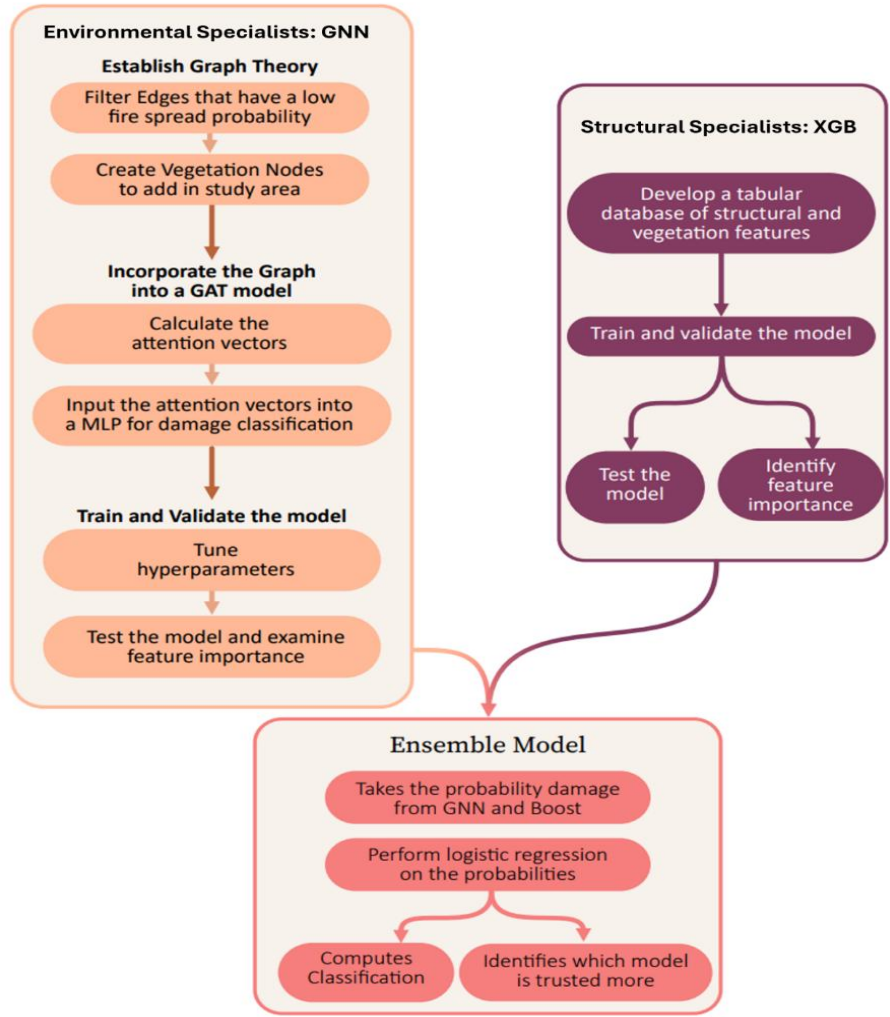

**Figure 2: Schematic of the proposed "Dual-Specialist" Ensemble Framework.** *The architecture bifurcates vulnerability assessment into two parallel processing streams to disentangle multi-scalar risk vectors.* **The Environmental Specialist (left):** *A graph neural network models neighborhood contagion dynamics by integrating a physics-informed topology—weighted by convection, radiation, and ember-spotting probabilities—with high-dimensional GAEF environmental embeddings via a graph attention (GAT) mechanism.* **The Structural Specialist (right).** *An XGBoost model isolates asset-level fragility by processing granular tabular features (e.g., eaves, siding, vents) using gradient boosting.* **Late Fusion (bottom):** *The independent damage probabilities from both specialist models serve as inputs for* **Logistic Regression Stacking***, which dynamically weights the contribution of landscape-level environmental pressure against building-level structural resistance to generate a holistic binary damage classification.*

For the GNN, there is a two-step process. The first step, detailed in section 4.1, focuses on graph construction. Similar to Chulahwat et al. (2022), it involves defining the network topology, specifying node features allocations, and calculating edge weights based on fire spread probabilities. This graph is then analyzed in the second step by a graph attention network (GAT). Section 4.2 explains the model architecture which yields a final binary prediction status (damaged or survived) for the test set. It is hypothesized that since the GNN has a high-dimensional environmental feature dataset, it will act as an environmental specialist that identifies buildings which are vulnerable to a fire due to local environmental features.

To provide complimentary insights, an XGB model is explored in parallel in section 4.3. Since the XGB model is dominated by the tabular structural features, despite having the Landfire features, it is hypothesized that it will act as a structural specialist that identifies buildings which are vulnerable to a fire due to their structural components. The workflow for the XGB model is rather simple as, unlike the GNN, it does not require any data construction. The tabular dataset of mostly structural features is used as an input for the XGB model. It divides the data in the training phase based on node purity. Once the model has made its predictions by splitting data, it can identify which feature was most useful for prediction.

The damage predictions for both the GNN and XGB model can enhance mitigation efforts by providing decision makers with knowledge related to the surrounding environment and structural robustness of the buildings. A logistic regression analysis compares the strengths of these models (Section 4.3). The inputs for this model are a softmax of the damage predictions from both models. These damage probabilities are compared to the ground truth label to compute the classification results. The coefficient weights of the model will highlight whether the GNN or XGB model had more influence for the respective events.

## 4.1 Graph theory

Graph networks consist of a series of nodes and edges which model complex interactions of the physical world and provide valuable information on how a phenomenon, such as wildfires, spread in heterogeneous study areas. While few studies leverage graph networks' capabilities for structure-level damage prediction in the WUI, existing prediction frameworks often rely solely on traditional network science metrics and omit deep learning techniques as seen in Chulahwat et al. (2022). The studies that do combine graph networks with deep learning assess fire spread mainly at large geographic scales rather than on the scale of individual buildings. This study bridges the existing methodological gap of wildfire preparedness at the structure level by applying proven graph theory approach to a deep learning network for assessing building damage.

To implement this framework, the network must be meticulously formulated to identify vulnerable structures before a fire event. This involves integrating house and vegetation nodes and calculating directed edges based on the probability of fire spread. First vegetation nodes are created and incorporated with the building nodes (Section 4.1.1). Then the edges are calculated based on the methodology developed by Chulahwat et al. (2022) (Section 4.1.2.) Critical low-impact connections are subsequently filtered out to focus the model on the most impactful propagation pathways. Figure 3 depicts the node types in the graph, how the GAEF embeddings are incorporated into all nodes, and the edge calculation process.

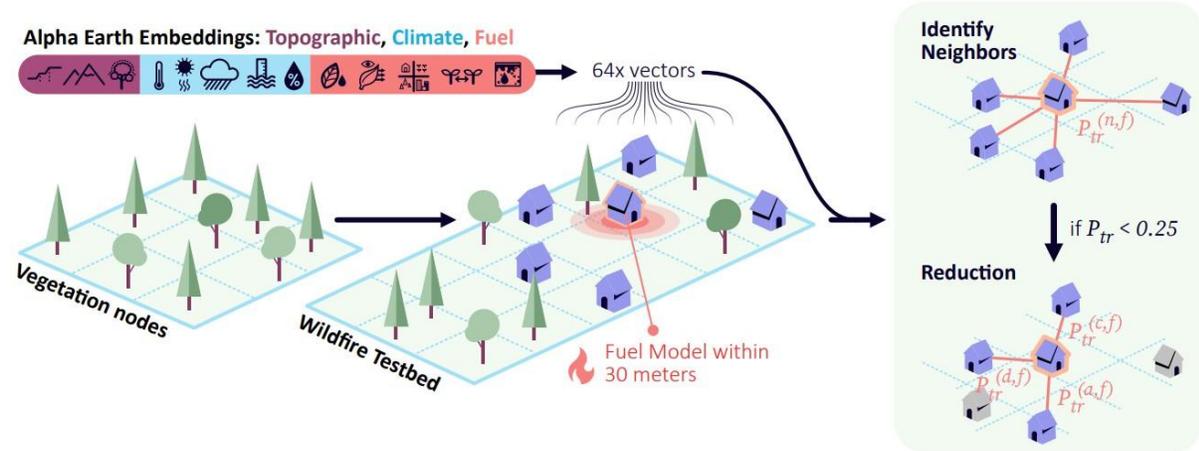

**Figure 3:** *Construction of the physics-informed contagion graph. The graph topology is engineered to simulate thermodynamic fire propagation through a three-stage pipeline:* **Node initialization (left):** *The WUI domain is discretized into heterogeneous building and vegetation nodes, each enriched with high-dimensional Google AlphaEarth Foundation embeddings and local topographic data.* **Mechanistic edge weighting (center):** *Directed edge weights $P_{tr}^{i,j}$ are computed via Monte Carlo simulation, aggregating the probabilities of convection, radiation, and ember spotting based on local fuel models.* **Topology construction and pruning (right):** *The initial graph topology was generated by considering all possible node pairs within a 200-meter radius. This created a graph with over 2 million edge possibilities. To isolate critical propagation pathways, edges with a cumulative transmission probability below 0.25 are filtered out, creating a sparse, optimized input graph that represents the most thermodynamically viable routes for fire spread.*

### 4.1.1 Development of nodes

In this study, nodes are represented by an individual structure or large patches of vegetation. To create the vegetation nodes, the LandFire raster data, where each pixel represents a 30-meter square area, is used. A discrete spatial resolution is established by generating a grid of points spaced 30 meters apart across the Eaton study area. These points are then assigned the corresponding fuel model, which is used to delete non-burnable nodes, ensuring all remaining nodes represent potentially burnable vegetation. Additionally, the fuel model will be used approximate variables for the edge calculation.

The data outlined in Section 3 are subsequently allocated to each node type, providing a rich feature set encompassing climate, weather, topography, structural vulnerability, and vegetation information. All nodes have GAEF embeddings and topographic features related to elevation and slope of the node. The structural features originate from the California Natural Resource Agency

Damage Inspection database and are encoded based on the vulnerability that component has to the fire. These features were allocated to building nodes with meaningful values, while vegetation nodes had values that the GNN would not consider useful in its analysis. Still, assigning a place-holder value to the vegetation nodes for building features was needed to ensure that the matrix dimensions of both node types were consistent and a necessary simplification of the model. Therefore, the total number of features for each node was 74, as there are 64 GAEF embeddings, two topographic features, and 8 structural features.

### 4.1.2 Development of edges

The edges' weights are calculated based on the methodology established by Chulahwat et al. (2022), which determines the total probability of fire spreading $P_{tr}^{i,j}$ from one node to another based on three primary modes of heat transfer: (1) convection, (2) radiation, and (3) ember spotting.

To calculate these methods of heat transfer mechanisms, the research used the fuel model for each node to assign a range of potential values for variables related to fire physics, such as flame height, flame temperature, and flame residence time. The susceptibility of target buildings was modeled using heat flux and flux time products according to the siding material of the building and the values in the d. Appendix A summarizes the approximate variables and their ranges.

To account for the inherent aleatoric uncertainty within these physical variables, the edge weights were computed using a Monte Carlo simulation. This method involved calculating each edge weight probability 100 times, with a random value taken from the defined range at every iteration for the necessary variables. The final edge weight assigned to the network was the average of these 100 simulated possibilities.

#### 4.1.2.1 Calculating the convection probability

Convection is the transfer of heat by the movement of fluids and corresponds to the ignition of an object due to the direct influence of flame. The probability of convection ($P_{conv}^{(i,j)}$) is defined by Eq. 1, which is unity if the distance ($d^{i,j}$) between the source node, $i$, and target node, $j$, are within threshold distance ($d_{conv}^{(i,j)}$), calculated in Equation 2, or zero otherwise.

$$P_{conv}^{(i,j)} = \begin{cases} 1 \text{ if } d^{i,j} \leq d_{conv}^{(i,j)} \\ 0 \text{ if } d^{i,j} > d_{conv}^{(i,j)} \end{cases} \quad \text{(Eq. 1)}$$

$$d_{conv}^{(i,j)} = F_{cc}^{(i,j)} * h_f^{(i)} * \tan(\theta_f) \quad \text{(Eq. 2)}$$

The term $d_{conv}^{(i,j)}$ is the product of the wind correlation coefficient, ($F_{cc}^{(i,j)}$; calculated in Eq. 3), height of the flame, $h_f^{(i)}$, and the tangent of the flame angle, $\theta_f$. Flame height was based on the fuel model assigned by the node. A detailed description of the approximation can be found in

Table 1A in Appendix A. To compute the flame angle, Eq. 4 Albini (1981) is used(Albini, 1981). The wind speed ($Ws^2$) is a constant of $22.2 \frac{m}{s^2}$ (80 kmph) since it was the worst-case Santa Ana scenario and direction of 225 degrees. These values come from an official Eaton "Early Insights" report from the Insurance Institute for Business & Home Safety (IBHS )(Insights, 2025). The value of gravity is $9.81 \frac{m}{s^2}$

$$F_{cc}^{(i,j)} = \begin{cases} \cos(|\phi^{(i,j)}-\theta|) \text{ if } |\phi^{(i,j)}-\theta|<90° \\ 0 \text{ if } |\phi^{(i,j)}-\theta|\geq 90° \end{cases} \quad \text{(Eq. 3)}$$

$$\theta_f = \arctan\left(\sqrt{\frac{3}{2} * \frac{Ws^2}{g*h_f^{(i)}}}\right) \quad \text{(Eq. 4)}$$

The wind correlation coefficient, $F_{cc}^{(i,j)}$, captures the effect of uncertainty in the wind direction by attaining a maximum value of unity when there is a perfect correlation between the wind direction, $\theta$, and the direction of edge from node $i$ to $j$ ($\phi^{(i,j)}$). $F_{cc}^{(i,j)}$ is computed by the piecewise function in Eq. 3, where it is the cosine of the absolute value of the difference between the edge direction and wind direction if that difference is less than 90 degrees or 0 otherwise.

### 4.1.2.2 Calculating radiation probability

When an object burns, it emits thermal radiation that can cause nearby objects to spontaneously combust. In the case of wildfires, sources of thermal radiation include buildings and vegetation. The methodology of calculating radiation in Chulahwat et al. (2022) is for a 3D structure and assumes a single household has multiple nodes. This study assumes a household is a single node. Therefore, the equations used in Chulahwat et al. (2022) are modified to match this assumption.

Eq. 5 calculates the probability of thermal radiation. If the minimum distances between the source and target node ($d_{min}^{(i,j)}$) are less than a 60-meter threshold ($d_{th}$), then the probability of radiation is set to be $P_l^{i,j}$ If the distance between two nodes is greater than 60-meters. $P_l^{i,j}$ is calculated by Eq. 6, which subtracts from 1 the cumulative density function that accounts for the time difference between the residence time of the flame ($t_r^i$) and time to ignition ($t_i^j$). The residence time of the flame is based on the fuel model (Table 1A, appendix A). The time of the flame is calculated by Eq. 7.

$$P_{rad}^{i,j} = \begin{cases} P_l^{i,j} \text{ if } d_{min}^{(i,j)} \leq d_{th} \\ 0 \text{ otherwise} \end{cases} \quad \text{(Eq. 5)}$$

$$P_l^{i,j} = \begin{cases} 1-F(t_r^i-t_i^j) \text{ if } t_r^i \geq t_i^j \\ 0 \text{ otherwise} \end{cases} \quad \text{(Eq. 6)}$$

$$t_l^{i,j} = \begin{cases} \overline{FTP^n(Q_l^{i,j}-Q_{cr}^n)^c} \text{ if } Q_l^{i,j} > Q_{cr}^n \\ 0 \text{ otherwise} \end{cases} \quad \text{(Eq. 7)}$$

In Eq. 7, $FTP^n$ is the flux time product of the material in the target direction. $Q_{cr}^n$ is the critical flux required for ignition of the target and $c$ is a constant based on the property of the target node. Finally, $n$ is the flux time product index. The aforementioned parameters are approximated based on the siding material of the building according to Table 2A in Appendix A.

$$Q_l^{i,j} = (\frac{A_j}{\pi * d_{min}^{(i,j)^2}} * \sigma * \varepsilon_i * T_f^4 - T_A^4) \quad \text{(Eq. 8)}$$

Eq. 8 calculates $Q_l^{i,j}$. $A_j$ is the area of the target node, and $d_{min}^{(i,j)}$ is the distance between two nodes. $\sigma$ is the Boltzmann constant, $5.67 \times 10^{-8}$, while $\varepsilon_i$ is the emissivity surface assumed to be 0.95. $T_f^4$ is the flame temperature which is approximated based on the source node's fuel model as seen in appendix A. $T_A^4$ is the event temperature which is assumed to be between 298.15 and 303.15 kelvin (25 °C and 30 °C) according to the IBHS Research for the Eaton fire (Insights, 2025).

### 4.1.2.3 Calculating the ember probability

Embers are one of the most predominant modes of fire propagation, as they travel further downwind than the actual fire front, thus enhancing the spread of the fire from node to node. This probability is computed by Eq. 9, where $g^{(i,j)}(.): \mathbb{R} \to [0,1]$ is the probability distribution function between node $i$ to $j$ given by a distribution function $S(i, d^{(i,j)}, v_w)$. The distribution is defined for each node pair interaction as a function of volume of the source node $V_n^i$, the distance between nodes $d^{(i,j)}$, and wind speed $v_w$. $P_{acc}^{(i,j)}$ is the probability of access for embers; the values are assumed to match those of Chulahwat et al. (2022) (Table 3A in Appendix A).

$$P_{ember}^{(i,j)} = g^{(i,j)} * P_{acc}^{(i,j)} * F_{cc}^{(i,j)} \quad \text{(Eq. 9)}$$

$$g^{(i,j)} = S(V_n^i, d^{(i,j)}, v_w) \quad \text{(Eq.10)}$$

### 4.1.2.4 Calculating total probability

Once all three mechanisms of heat transfer are calculated, the unity between these probabilities is computed as shown in Eq. 11.

$$P_{tr}^{i,j} = \left(P_{conv}^{(i,j)} \cup P_{rad}^{i,j} \cup P_{ember}^{(i,j)}\right) \quad \text{(Eq. 11)}$$

Each of the three probabilities were computed 100 times, as several variables needed to be approximated based on the fuel model and material of the building. The average value of each probability was assigned as the edge weight. Edges with value of $P_{tr}^{i,j}$ less than 0.25 were filtered out similarly to the manner of Chulahwat et al. (2022). Once the node features and edge weights are assigned in the graph, they are used as an input for the GNN as described in section 4.2.

## 4.2 Deep learning graph neural network

Figure 4 illustrates the DL workflow. Once the graph is created, as described in section 4.1, it is used as an input for the GNN architecture, which deploys a graph attention network layer. The GAT accepts a vector based on the 74 node features from the graph input for feature transformation that translates the heterogenous risk inputs into a specialized context-aware vector for damage classification. These context-aware vectors are then used as an input for a multilayer perceptron (MLP) layer.

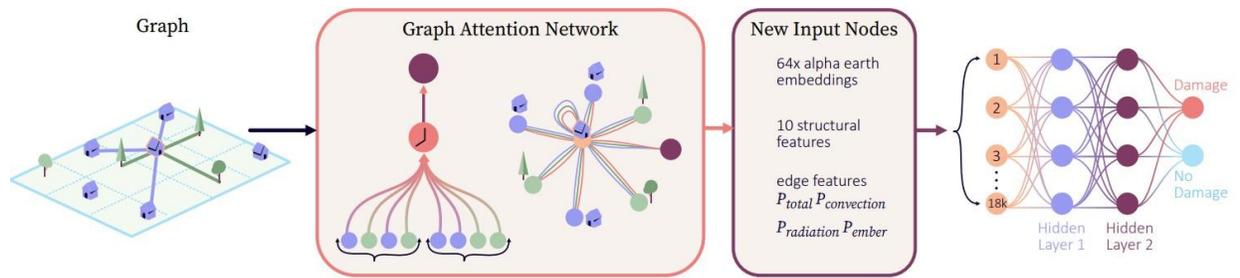

**Figure 4:** *Deep learning architecture of the environmental specialist (GNN). The architecture transforms the physics-informed graph into a predictive risk assessment. The input graph is processed by a **graph attention layer**, which uses a multi-head attention mechanism to dynamically aggregate features from neighboring nodes. This transformation projects the initial 74-dimensional raw feature vector (structural + environmental) into a dense **256-dimensional context-aware embedding** (64 dimensions × 4 heads). This latent representation, which now encodes both local attributes and neighborhood contagion pressure, is passed through a multi-layer perceptron classification head to output the binary probability of damage.*

The GAT architecture leverages masked self-attention layers to transform complex, non-grid-like data into a predictive risk vector. This approach allows the network to assign varying levels of importance to neighboring nodes based on the respective features of the nodes as well as the edge weights. The GAT layer accepts the full input feature set, $h = \{h_1, h_2 \ldots \ldots h_N\}$ where the total number of nodes, $N$, is 48,470 (18,422 building nodes and 30,048 vegetation nodes) for the Eaton study area. Each node contains the 74-feature set which includes 64 GAFE, eight structural attributes, and two topographic features.

To initialize the learning process and prepare features for attention, the GAT layer first applies a shared learnable linear transformation that is parameterized by the weight matrix W. Additionally, the GAT incorporates Edge features, which represent the fire spread probability $\left(P_{tr}^{i,j}, P_{conv}^{(i,j)}, P_{Rad}^{(i,j)}, P_{ember}^{(i,j)}\right)$ as a distinct linearly transformed input feature within the calculation of the original attention coefficient. The Edge features are linearly transformed by a separate dedicated weight matrix Θ. The attention coefficient $e_{ij}$ is computed by fusing the transformed Node features and Edge weights as seen in Eq. 12. To capture the non-linear interactions, a LeakyReLU activation function is applied to the concatenation of the source node, target node, and Edge feature vectors.

$$e_{ij} = LeakyReLU(\vec{a^T}[W\vec{h_i} \| W\vec{h_j} \| \Theta_e e_{ij}]) \quad \text{(Eq. 12)}$$

As shown in Figure 5, the GAT's self-attention mechanism is analogous to query-key value framework of transformers. The central Node's features $W\vec{h_i}$ act as the query, while the neighboring node's features serve as the key and value. This framework allows the network to contextually learn which neighbors contribute most significantly to the central node's wildfire risk.

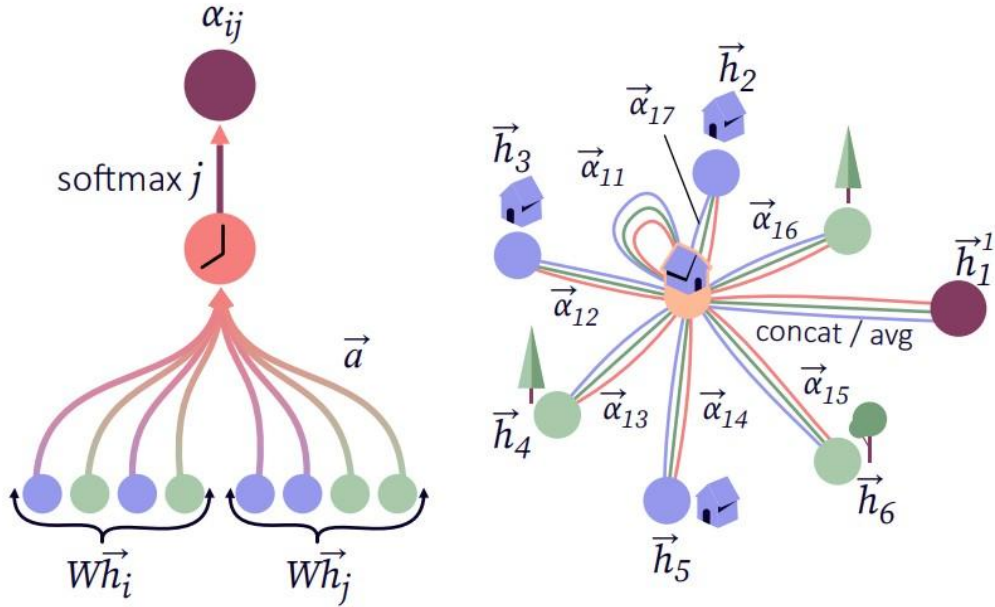

**Figure 5:** *The attention mechanism. Adapted from* (Veličković et al., 2017), *the schematic shows that the weights of the observed node and neighboring node being used to calculate the attention score $\alpha_{ij}$. These alphas will then be used as inputs for the transformed vector $h_i'$*

Additionally, Figure 5 shows how connections between nodes vary. It is critical to keep the same scale throughout the analysis; therefore, the attention coefficient needs to be normalized by using the Softmax function (Eq. 13).

$$\alpha_{ij} = \text{softmax}(e_{ij}) = \frac{\exp(e_{ij})}{\sum_{k \in N_i} \exp(e_{ik})} \quad \text{(Eq. 13)}$$

To improve the stability of the learning process, multihead attention is employed with K=4 independent attention heads. The output of these heads is concatenated to preserve the diversity of the learned features. The final aggregated vector, $h'_i$ is calculated as an average of the results from K independent attention heads, $\alpha^k_{ij}$, where σ is the final activation function (Eq. 14):

$$h'_i = \Big\|_{k=1}^{K} \Big(\sigma\Big(\sum_{j \in N_i} \alpha^k_{ij} W^K \vec{h_j}\Big)\Big) \quad \text{(Eq. 14)}$$

The hidden dimensions for each of the K=4 attention heads (( set to 64. By concatenating the four heads, the GAT layer produces a 256-dimensional context aware output vector (64x4) for each node. This vector, which was originally the 74 feature vectors entered earlier, now contains information regarding neighboring node features and edge-weight context. This 256-output vector serves as the input for the MLP (Figure 4). The MLP has two hidden layers with the ReLU activation function to perform the binary classification task.

## 4.3 Machine learning methods

In addition to deep learning, this research explores the application of ML for wildfire preparedness at the building level. The GNN can process high-dimensional data exceptionally well, and decision trees, such as XGB, excel at assessing tabular data. Therefore, both models in this study can provide insights on complex environmental traits (from the GNN model) and structural features (from the XGB model) that could make a building vulnerable to wildfire. Another ML method used in this study is logistic regression. This allows the research to combine both model's damage probabilities in a combined analysis.

### 4.3.1 XGBoost Decision tree method

At its core, building a decision tree involves selecting the best feature for data splitting at each node based on the Gini impurity for this classification task. The process iterates until the tree reaches a specified depth or node purity.

This provides an interpretable model as insights such as feature importance is discovered by identifying the parameter that is a key drive when making the splits. Vanilla decision trees are prone to overfitting which can hinder its generalization to new data; therefore, the XGBoost Classifier is selected. This is an optimized version of gradient boosting. It is characterized by building a sequential ensemble of decision trees where each subsequent tree is trained to correct the errors of its predecessor, minimizing loss via gradient descent. This is particularly advantageous when working with the tabular datasets as it is effective at capturing nuanced interaction effects among features, which is crucial for classification tasks. In the context of this research, it allows the model to distinguish between multiple structural features that could cause a building to be vulnerable to wildfire.

### 4.3.2 Logistic regression method

The main component of the GNN would be the GAEF embeddings since they both provide rich environmental feature data for the model's predictions. The main focus of the XGB model is structural building features. These two models complement each other's insights nicely, as the two most critical components in wildfire damage assessment are the environmental and building features. This provides an opportunity for this research to create an ensemble model. The regression model can be described in Eq. 15:

$$\ln\left(\frac{P_{Damage}}{1-P_{Damage}}\right) = \beta_0 + \beta_{GNN}x_{GNN} + \beta_{XGB}x_{XGB} \qquad (Eq.\ 15)$$

A logistic regression model is implemented that takes the damage probabilities from each model. By adopting a late fusion approach, this research mitigates the individual GNN and XGB model's prediction errors and quantifies the distinct contribution of environmental and structural factors to the final damage classification as the damage probabilities for both models will be inputs for the logistic regression that as the same ground truth damage label.

## 5. Results and Discussion

This analysis is structured around quantifying the contribution of each specialist model. Section 5.1 describes the classification performance and determines the extent that environmental, structural, and graph topology features influence the GNN's prediction. Section 5.2 details the XGBoost classification metrics and feature importance. Finally, the role of the ensemble framework is summarized, in section 5.3, by the logistic regression model, which quantifies the value of both models' damage prediction probabilities.

### 5.1 GNN results

A graphical assessment combined with deep learning for wildfire damage prediction is a novel approach due to its ability to consider high-dimensional data, community layout, and ability to capture general trends for transferability. To achieve high-fidelity damage classification, this study uses a GNN that leverages GAT layers within a HetroConv framework for feature processing, feeding the vectorized node embeddings into a multilayer perceptron head for binary classification. The model employs a single GAT layer utilizing that concatenate outputs from four attention heads. The overall network and training process are governed by the following hyperparameters:

- Network architecture: The GNN uses one graph layer with four attention heads, a hidden dimension of 64, and a two-layer MLP classification head.
- GAT layer parameters: Dropout is set at 0.05, there is no self-loop, and the outputs are concatenated.
- Optimization: The Adam optimizer is used with an initial learning rate of $5x10^{-4}$. The weight decay with L2 regularization is set at $1x10^{-6}$. Then a final dropout of 0.1 is applied to the MLP heads.
- Training: The model is trained for 1,000 epochs with a patience of 100 for early stopping and an optional learning rate scheduler that decays by 0.9 every 50 epochs.

The model was able to achieve an F1 score of 0.82 for survived structures and 0.86 for damaged structures. The precision and recall for damaged structures were 0.87 and 0.86, respectively, demonstrating the model's ability to predict damaged structures evident in the confusion matrix in Figure 6a as 1,354 out of 1,554 damaged structures were correctly classified. For survived structures, the precision was 0.82; recall was 0.83. The slightly lower precision for survived structures (0.82) indicates a conservative prediction bias, as the model mistakenly flags a small percentage of surviving homes as damaged (false positives). The confusion matrix shows that 203 out of 1,185 structures were classified as damaged, while the remaining 982 were correctly classified.

**Table 1: GNN classification metrics**

|  | Survived | Damaged |
|---|---|---|
| **Precision** | 0.82 | 0.87 |
| **Recall** | 0.83 | 0.86 |
| **F1-Score** | 0.82 | 0.86 |
|  | Overall | |
| **Accuracy** | 84.67% | |
| **F1 Score** | 0.844 | |

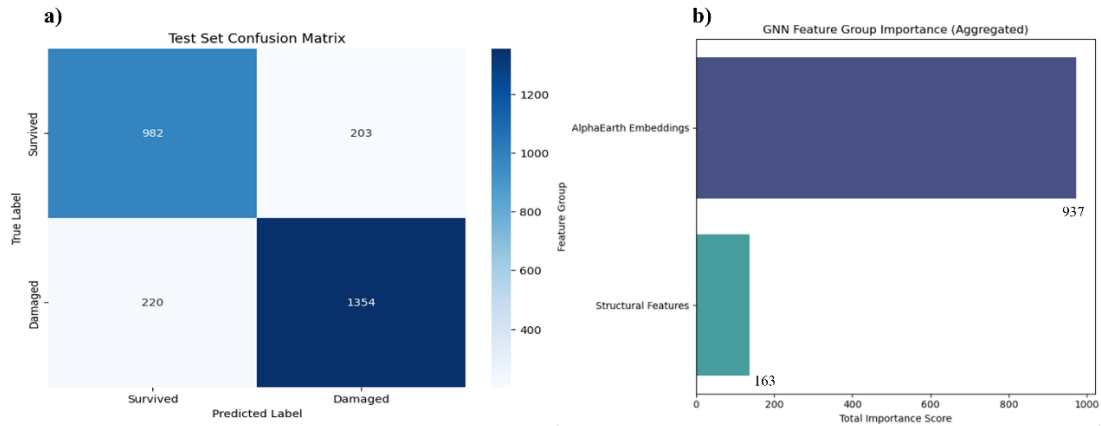

**Figure 6:** *GNN predictive performance and feature attribution. (a) Test set confusion matrix: The model demonstrates robust sensitivity to destruction, correctly classifying 1,354 damaged structures (true positives). The 203 false positives (survived homes predicted as damaged) represent* **latent risk**—*properties located in high-connectivity danger zones that survived due to stochastic factors (Note: Total damaged count is 1,574). (b) Aggregated feature importance: Internal attention weights reveal the hierarchy of the Environmental Specialist. The* **GAEF environmental embeddings** *(Score: 937) overwhelmingly dominate the tabular structural features (Score: 163), quantitatively confirming that the GNN prioritizes landscape-scale exposure over building-level fragility.*

The model's internal attention weights were extracted and aggregated by feature type to quantify the relative contribution of structural attributes versus geospatial embeddings. Figure 6b shows that the GAEF embeddings have a group importance score of 937, while the structural features have a score of 163. This difference highlights the critical importance of incorporating rich, location-based environmental data into structural damage models. As a network-aware model, the GNN's architecture is suited to process these high-dimensional feature vectors and to extract predictive spatial context. This integration is necessary because successful wildfire risk reduction requires a holistic understanding that fully integrates the wildland's characteristics and the built environment's features.

To validate the physical realism of the GNN's predictions, we analyzed the centrality measures of test nodes relative to their predicted outcomes. Weighted in-degree centrality, representing the cumulative ignition pressure from neighboring nodes, was calculated for the full graph. The analysis revealed that false positives (homes predicted to be destroyed but which survived) had the highest average centrality scores (0.322) of any group, significantly higher than true positives (0.184) or true negatives (0.231). This result shows that the GNN identified survived homes as damaged due to their high connectivity to hazardous neighbors.

**Table 2: Degree Centrality**

| Outcome | Degree centrality (mean, standard deviation) | Eigenvector (mean, standard deviation) | Number of samples |
|---|---|---|---|
| **False negative** | 0.239 | $7.5 * 10^{-5}$ | 220 |
| **False positive** | 0.322 | $4.3 * 10^{-5}$ | 203 |
| **True negative** | 0.231 | $5.9 * 10^{-5}$ | 982 |
| **True positive** | 0.184 | $2.1 * 10^{-5}$ | 1354 |

For other fire events, graph structure has proven to be an important predictive consideration, exemplified by the findings of Chulahwat et al. (2022), who achieved a maximum of 64% accuracy in damage prediction based purely on graph topology. The findings from this research both complement and advance that methodology: they demonstrate that incorporating rich, informative node features in a higher-dimension space provides more robust predictions, yet this input dominates the model's decision-making, rendering the isolated graph structure a minor factor for this specific fire event. The GNN novelty lies in its ability to effectively incorporate both node and graph structure. The graph structure is embedded into the prediction by utilizing the graph attention network layer, which computes attention coefficients that dynamically fuse edge features $\left(P_{tr}^{i,j}, P_{conv}^{(i,j)}, P_{Rad}^{(i,j)}, P_{ember}^{(i,j)}\right)$ with neighboring node data. Despite the observed dominance of the environmental features, it is still critical to consider graph structure, as it acts as a generalization safeguard for future fire events. If omitted, foundational knowledge may be overlooked, potentially leading to poor decision-making. These results demonstrate that the GNN successfully combines advanced environmental embeddings with structural features to provide a highly accurate and balanced risk assessment, proving the effectiveness of this hybrid deep learning approach for pre-disaster wildfire vulnerability analysis. The dominance of these environmental features reinforces the immediate need for coupling wildland and built-environment mitigation strategies. By identifying vulnerable structures based on this rich environmental context, the GNN can directly aid in orchestrating effective mitigation, such as strategically placing firebreaks and reinforcing households with wildfire-resistant design.

## 5.2 XGBoost Results

To complement the findings from the GNN, an XGBoost classifier was used to serve as the structural specialist model. The advantage of using a tree-based model is their ability to efficiently process both categorical structural features and numerical environmental data. The model was trained using the built in XGBoost classifier with 300 boosting trees, a maximum depth of 6, and a 0.1 learning rate.

The model's performance on the test set (which shared the same stratification as the GNN) demonstrated high predictive capability, achieving an overall accuracy of 88%. The F1 score reached 0.87 for survived structures and 0.89 for damaged structures. Specifically, the model exhibited superior confidence in identifying structural risk, achieving a precision of 0.93 for the damaged category. Conversely, the model effectively identified safe homes, evidenced by a strong recall score of 0.91 for the Survived class.

**Table 3: XGB model classification results**

|  | Survived | Damaged |
|---|---|---|
| **Precision** | 0.83 | 0.93 |
| **Recall** | 0.91 | 0.86 |
| **F1-Score** | 0.87 | 0.89 |
|  | Overall | |
| **Accuracy** | 88% | |
| **F1 Score** | 0.88 | |
| **AUC** | 0.915 | |

When observing feature importance, the Eaves feature was determined to be the overwhelmingly dominant predictor, accounting for 0.666 of the model's total gain. This finding directly aligns with established structural vulnerability research by Noumeur et al. (2025), which identified vent screens, eaves, and roof material as critical structural components. The rationale is that these features are ingress points for embers.

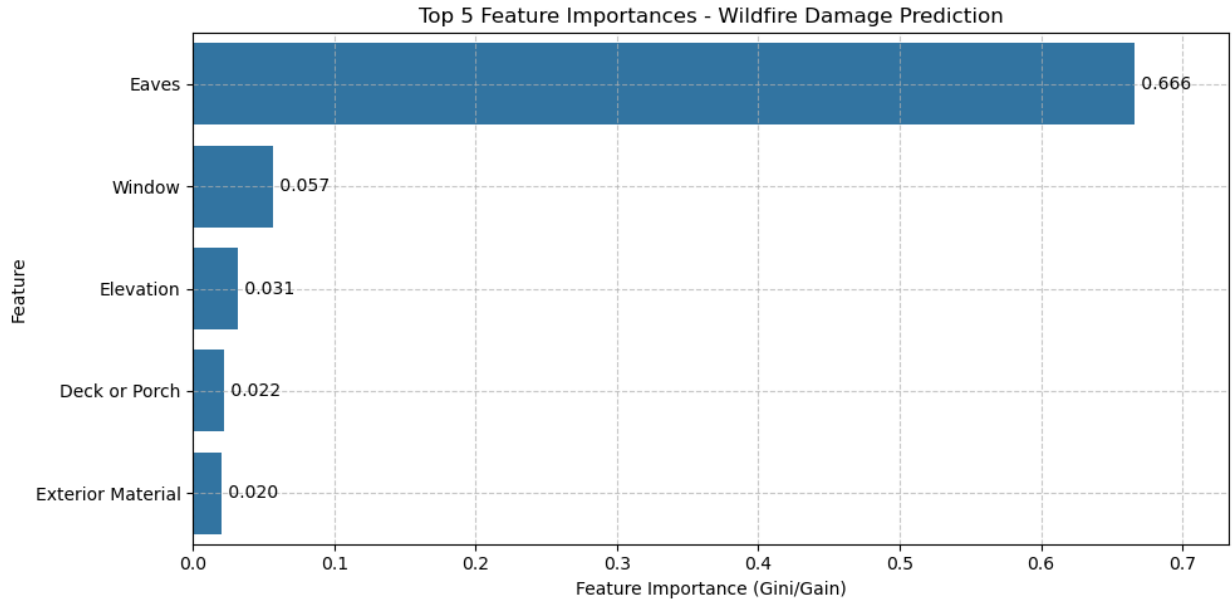

**Figure 7:** *The XGB model shows that the Eaves feature was the most important feature during the Eaton fire event. eaves can act as an entry barrier from fire embers.*

When comparing the two specialist models, the XGB model offers an advantage in interpreting tabular data and providing actionable building-specific guidance; however, the XGB model is limited in its ability to generalize or effectively process the high-dimensional GAEF features. Conversely, the GNN excels at fusing high-dimensional data and modeling spatial contagion but is less effective at highlighting specific structural components. This complementarity justifies the ensemble framework, which combines the structural precision of the XGB model with the superior generalizability and environmental awareness of the GNN

## 5.3 Logistic regression analysis

The two specialist models, GNN and XGB, were developed to provide unique insights into fire risk. The GNN focuses on hazard and contagion by leveraging high-dimensional environmental and topographic features, while the XGB model acts as a structural specialist, prioritizing localized building features. This complementary design ensures that the final ensemble captures the full complexity of WUI risk, bridging the gap between large-scale environmental exposure and individual asset vulnerability.

To demonstrate the harmony between both models, their damage probability predictions are used as inputs for a logistic regression analysis. The results of this experiment can be seen in Table 4. It shows that the F1 score for structures that survived the fire was 0.82; for damaged structures, the scores was 0.87, with 85% accuracy.

Additionally, the logistic regression model has a higher coefficient weight for the GNN probabilities (5.39) compared to the coefficient weight of the XGBoost probabilities (0.076). This shows that for the Eaton fire the environmental features were more dominant when estimating damage potential. The dominance of environmental features may not be the case in future fire events; therefore, it is critical to consider multiple aspects of potential vulnerabilities such as structural vulnerabilities and topographic layout vulnerabilities.

**Table 4: Logistic regression classification results**

|  | Survived | Damaged |
|---|---|---|
| **Precision** | 0.82 | 0.87 |
| **Recall** | 0.82 | 0.87 |
| **F1-Score** | 0.82 | 0.87 |
|  | **Overall** | |
| **Accuracy** | 85% | |
| **F1 Score** | 0.84 | |

The GNN considers environmental features and topographic layout of the community, while the XGBoost model focuses mostly on structural features. By combining the two models and interpreting their results, decision makers can develop robust vulnerability plans. For example, Figure 8 shows the test data divided into four vulnerability categories. If the probability of the GNN prediction was greater than 0.5 and XGB was below 0.5, these structures are considered vulnerable due to environmental features. One mitigation strategy could be examining the defensible space of these areas. Conversely, if the XGB model had a probability greater than 0.5 and GNN probability less than 0.5 these structures are vulnerable due to building features. One mitigation strategy can be examining the eaves, windows, vents and patio covers to ensure these structures are safe from embers. If both probabilities are greater than 0.5, they are prone to environmental factors and structural factors. Therefore, these structures should combine both mitigation strategies. If both probabilities are below 0.5, they are marked safe.

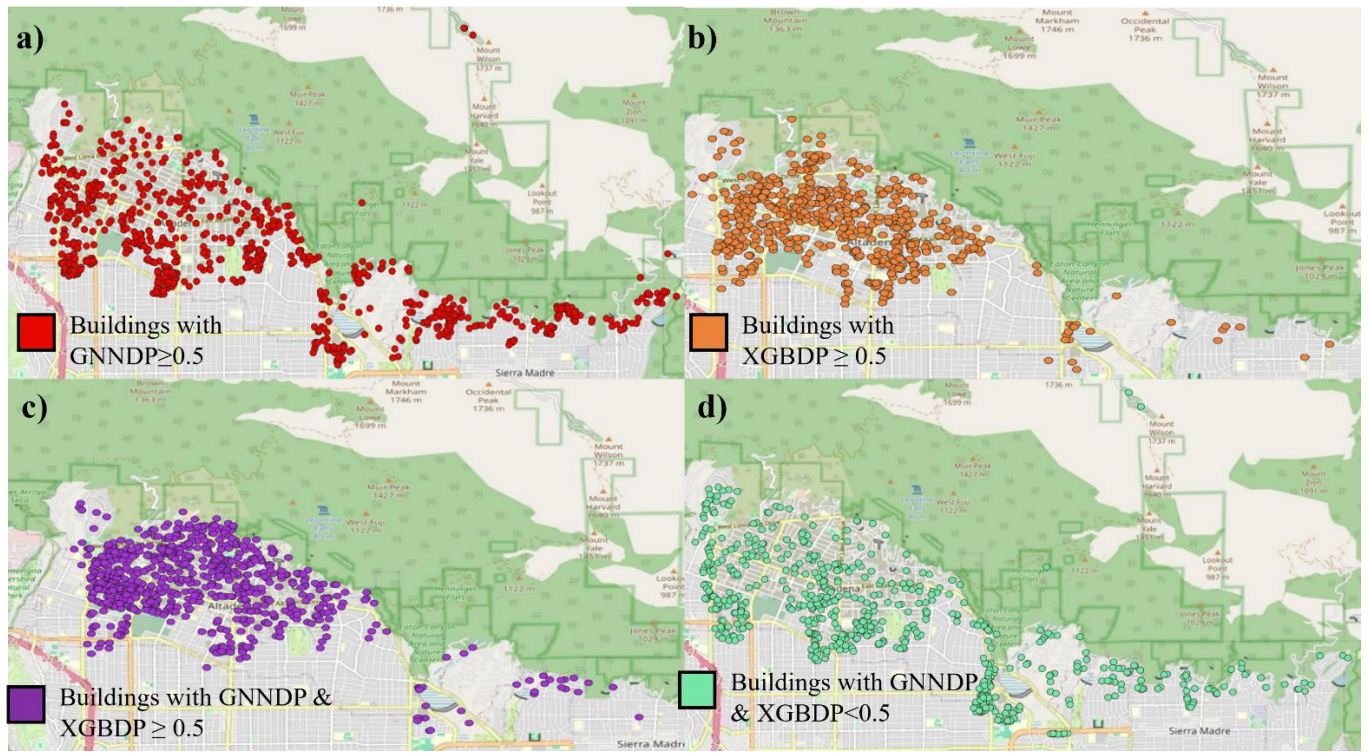

**Figure 8:** *The damage probability of the test-set buildings. 8a Locations of buildings vulnerable to the fire due to the local environment, as the GNN damage probability is greater or equal to 0.5. 8b Locations of buildings vulnerable due to structural features as the XGB damage probability is greater or equal to 0.5. 8c Buildings vulnerable to fire due to both environmental and structural features. 8d Buildings not vulnerable to fire.*

## 6.0 Conclusion

This paper presents a methodological framework to enhance wildfire preparedness accomplished by utilizing a GNN that incorporates high-dimensional environmental data along with structural features of the building. The GNN achieved an F1 score of 0.82 for survived buildings and 0.86 for destroyed buildings. The main contribution of these scores was the Google AlphaEarth Foundation embeddings, as the group importance score was 937, compared to a structural feature score of 163. To compliment these results, the research examines an XGB model that focuses on the structural features of the buildings. The XGB model had an F1 score of 0.87 for survived and 0.89 for destroyed buildings. The most important structural feature for this model were Eaves. This is due to Eaves being a critical entry point for embers during a wildfire. The results of the GNN are slightly lower than those of the XGB; however, both models complement each other's findings, as seen in the logistic regression analysis and Figure 8. The GNN model focused on environmental features, while the XGB focused on structural features. In the logistic regression analysis, the GNN had more weight, but including the XGB model's damage probabilities improved 12 classifications when compared to assessing purely the GNN results. Figure 8 shows buildings vulnerable due to environmental features, structural components, and both. This knowledge can enhance planning and preparedness by giving decision makers critical information on what measures to take to enhance their community's resilience.

Accordingly, this study advances the theoretical framework of wildfire–urban interface by the GNN's heavy prioritization of geospatial embeddings (937) over structural features (163). This dichotomy challenges the sufficiency of parcel-level assessments. Although the XGBoost model validated established engineering principles by identifying as the primary ingress vector (aligning with Numerus et al., 2025), the ensemble's substantial weighting of the GNN (coefficient: 5.39) implies that during extreme propagation events, neighborhood-scale environmental conditions often override individual structural defenses. Furthermore, the observation that false positive predictions possessed the highest degree centrality scores reveals a novel class of latent risk—properties that likely survived due to stochastic factors rather than inherent resilience. This finding suggests that future disaster analytics must pivot from binary classification toward probabilistic "siege analysis" that quantifies when community-level exposure renders property-level mitigation insufficient.

Translating this ensemble framework into practice requires a three-phase "diagnostic triage" strategy that moves stakeholders from reactive suppression to proactive hardening. Phase I (data Proxy Calibration) overcomes the barrier of sparse pre-event structural data by integrating municipal tax assessor records and computer-vision analysis of street-view imagery to populate the granular inputs (e.g., eaves, vents) required by the structural specialist model. Phase II (targeted vulnerability segmentation) operationalizes the ensemble's split-logic to optimize resource allocation: urban planners utilize the GNN's environmental pressure maps to designate Community-Scale Intervention Zones for fuel breaks in high-contagion neighborhoods, while fire marshals leverage the XGB model's structural fragility scores to target homeowner retrofit grants toward properties in "structural-risk dominant" clusters. Phase III (impact verification) establishes the framework for measuring success beyond static prediction accuracy. While barriers such as data privacy constraints and the computational overhead of processing high-dimensional embeddings persist, real-world application should be evaluated by the mitigation conversion rate—the quantifiable percentage of high-risk nodes successfully migrated to a lower vulnerability quadrant through targeted intervention. This metric provides insurers and policymakers with a verifiable return on resilience investment.

The contribution of this research is the methodological framework of (1) combing graph theory with deep learning to create a model that can enhance wildfire preparedness and (2) demonstrating the potential for an ensemble workflow that combines the strengths of two mathematical models to holistically assess building vulnerability. Future research can build off this study by exploring other ensemble applications in different domains such as flooding. One limitation in this study is the binary classification. While it is more critical to identify damaged buildings versus survived buildings in the preparedness phase of a wildfire, having granular insight of building damage could still be useful and give practitioners more flexibility during the planning phase. Studies that have explored granular applications of building damage have acknowledged the imbalance issue. This issue provides avenues for future research into enhance these models to handle data imbalance, such as generating synthetic data that captures the minor, major, and affected categories robustly, thus giving the model more data from which to learn. Another limitation of this research is that the study only assesses the Eaton Fire in California as its case study. Including more events and extending the GNN to predict edge weights based on node features could improve the generalization, and thus, the transferability of the GNN model.. This will bypass the need to calculate the edge weights in future events and improve implementation speed of the model. This research presents a novel workflow to robustly identify environmental and structural vulnerabilities of buildings during the preparedness phase of a wildfire. This will enhance resource allocation and improve mitigation efforts in areas that are prone to wildfires as urban expansion continues into vulnerable land.

# Appendix A

**Table 1A: Fuel Model flame variable ranges from Scott and Burgan 2005**

| Fuel Model category | Range of values for flame length (meters) | Range of values for residence time (seconds) | Flame temperature (Kelvin) |
|---|---|---|---|
| **Very Low** | (0,0.3048) | (20,40) | (750,850) |
| **Low** | (0.3048,1.219) | (40,80) | (85tgh0,1000) |
| **Moderate** | (1.219,2.438) | (120,300) | (1000,1500) |
| **High** | (2.348,3.658) | (300,600) | (1150,1250) |
| **Very High** | (3.658, 7.620) | (600,1200) | (1250,1350) |
| **Extreme** | (7.620,15.24) | (1200,2400) | (1350,1450) |
| **Non-Burnable** | (0,0) | (0,0) | (293,293) |

**Table 2A: Material values adapted from USNRC**

| Material | Heat flux ($Q_{cr}^n$) (kW/m2) | Flux time product (kW-sec/m2) | Flux time product index n |
|---|---|---|---|
| **Wood** | (8.5,13.7) | (5130,6164) | (1.5,1.53) |
| **Vinyl** | (14,16) | (4800,5400) | (1.45,1.55) |
| **Stucco Brick Cement** | (10,000, 10,000) | (20000,20000) | (1.0,1.0) |
| **Vegetation** | (10,15) | (300,1200) | (1.0,1.25) |
| **Unknown** | (30, 40) | (6000,8000) | (1.5,1.5) |

**Table 3A: Material values similar to Chulahwat et al. (2022)**

| Structural Feature | Type | Value |
|---|---|---|
| **Deck or porch on grade** | (Composite, masonry or concrete, wood, none) | (0.3,0.3,2.7,2) |
| **Eaves** | (Composite, masonry or concrete, wood, none) | (0.3,0.3,2.7,2) |
| **Roof** | (Wood, composite, tile, concrete, metal, asphalt, other) | (4.1,0.7,0.3,0.3,0.3,0.7,1) |
| **Vent Screen** | (Mesh < 4mm, Mesh> 4mm, No vents, no screen) | (0.7,1.2,1.1,1.5) |
| **Fence** | (Combustible, non-combustible, none) | (1.8,1.1,0.7) |
| **Window** | (Multi-pane, single-pane) | (0.4,3.0) |

# Appendix B

**Table 1B: Feature allocation and description of node data.**

|  | **Building Node** | **Vegetation Node** |
|---|---|---|
| **Alpha Earth embeddings** | The 64 vector embeddings are mapped to the respective latitude and longitude of the building node. | The 64 vector embeddings are mapped to the respective latitude and longitude of the vegetation node |
| **Structural features** | A structural vulnerability score is assigned to the material of the roof, building siding, eaves, how many windows the building has, if the deck is present or not and is flammable. This score is based on (Meldrum et al., 2022) | The structural features of the vegetation node are omitted |
| **Vegetation features** | The most conservative flammable vegetation within 30 meters of a house is assigned to provide insight on flame length, residence time, temperature of the fire | The flame length, residence time, temperature of the fire is approximated according to the fuel model category |
| **Topographic features** | The area of the building was computed based on Microsoft building footprint, slope and elevation were taken from the DEM database | The area was computed by multiplying the 30mx30m grid cell siding, and slope and elevation were taken from the DEM database |
| **Total number of features** | 64 alpha earth embeddings<br><br>eight structural features<br><br>two topographic features | 64 alpha earth embeddings<br><br>eight structural features, but are set to zero. This was to make the matrix math consistent<br><br>two topographic features |
| **Volume approximation input for equation 10** | Microsoft building footprint data only has the area and to obtain the volume, the height of the buildings must be approximated.<br><br>The research grouped together building types and allocated an approximated average height value for each type. This height was multiplied by the respective area of the building to obtain the volume used in equation 10. | The vegetation volume was approximated based on depth of the fuel model multiplied by the 30m x 30m area |

# References

Albini, F. A. (1981). *A Model for the Wind-Blown Flame from a Line Fire\**. *174*, 155–174.

Alguliyev, R., Aliguliyev, R., & Yusifov, F. (2021). Graph modelling for tracking the COVID-19 pandemic spread. *Infectious Disease Modelling*, *6*, 112–122. https://doi.org/10.1016/j.idm.2020.12.002

Andrews, P. L. (2006). *BehavePlus FIRE MODELING SYSTEM : PAST , PRESENT , AND FUTURE \* Corresponding author address : Patricia L . Andrews , Fire Sciences Lab , 5775 Highway 10 W , Missoula , MT*.

Boroujeni, S. P. H., Razi, A., Khoshdel, S., Afghah, F., Coen, J. L., O'Neill, L., Fule, P., Watts, A., Kokolakis, N. M. T., & Vamvoudakis, K. G. (2024). A comprehensive survey of research towards AI-enabled unmanned aerial systems in pre-, active-, and post-wildfire management. *Information Fusion*, *108*(March), 102369. https://doi.org/10.1016/j.inffus.2024.102369

Brown, C. F., Kazmierski, M. R., Pasquarella, V. J., Rucklidge, W. J., Samsikova, M., Zhang, C., Shelhamer, E., Lahera, E., Wiles, O., Ilyushchenko, S., Gorelick, N., Zhang, L. L., Alj, S., Schechter, E., Askay, S., Guinan, O., Moore, R., Boukouvalas, A., & Kohli, P. (2025). *AlphaEarth Foundations: An embedding field model for accurate and efficient global mapping from sparse label data*. http://arxiv.org/abs/2507.22291

Cao, Y., Zhou, X., Yu, Y., Rao, S., Wu, Y., Li, C., & Zhu, Z. (2024). *Forest Fire Prediction Based on Time Series Networks and Remote Sensing Images*. 1–26.

Cardil, A., Monedero, S., Schag, G., de-Miguel, S., Tapia, M., Stoof, C. R., Silva, C. A., Mohan, M., Cardil, A., & Ramirez, J. (2021). Fire behavior modeling for operational decision-making. *Current Opinion in Environmental Science and Health*, *23*. https://doi.org/10.1016/j.coesh.2021.100291

Chulahwat, A., Mahmoud, H., Monedero, S., Diez Vizcaíno, F. J., Ramirez, J., Buckley, D., & Forradellas, A. C. (2022). Integrated graph measures reveal survival likelihood for buildings in wildfire events. *Scientific Reports*, *12*(1), 1–17. https://doi.org/10.1038/s41598-022-19875-1

Clare, J., Garis, L., Plecas, D., & Jennings, C. (2012). Reduced frequency and severity of residential fires following delivery of fire prevention education by on-duty fire fighters: Cluster randomized controlled study. *Journal of Safety Research*, *43*(2), 123–128.

Dabrowski, J. J., Pagendam, D. E., Hilton, J., Sanderson, C., MacKinlay, D., Huston, C., Bolt, A., & Kuhnert, P. (2023). Bayesian Physics Informed Neural Networks for data assimilation and spatio-temporal modelling of wildfires. *Spatial Statistics*, *55*. https://doi.org/10.1016/j.spasta.2023.100746

DaCosta, M., Krinsley, J., & Abelson, B. (2015). Optimizing local smoke alarm inspections with federal data. *Bloomberg Data for Good Exchange*, *119*, 379–7112.

Dargin, J., Berk, A., & Mostafavi, A. (2020). Assessment of household-level food-energy-water nexus vulnerability during disasters. *Sustainable Cities and Society*, *62*(June), 102366.

https://doi.org/10.1016/j.scs.2020.102366

Dong, S., Esmalian, A., Farahmand, H., & Mostafavi, A. (2020). An integrated physical-social analysis of disrupted access to critical facilities and community service-loss tolerance in urban flooding. *Computers, Environment and Urban Systems*, *80*, 101443.

Dong, S., Gao, X., Mostafavi, A., & Gao, J. (2022). Modest flooding can trigger catastrophic road network collapse due to compound failure. *Communications Earth \& Environment*, *3*(1), 38.

Esparza, M., Esmalian, A., Dong, S., & Mostafavi, A. (2021). Examining spatial clusters for identifying risk hotspots of communities susceptible to flood-induced transportation disruptions. In *Computing in civil engineering 2021* (pp. 482–489). American Society of Civil Engineers.

Esparza, M., Farahmand, H., Liu, X., & Mostafav, A. (2024). Enhancing inundation monitoring of road networks using crowdsourced flood reports. *Urban Informatics*, *3*(1). https://doi.org/10.1007/s44212-024-00055-7

Esparza, M., Gupta, A., Mostafavi, A., Yin, K., & Xiao, Y. (2025). Automated wildfire damage assessment from multi view ground level imagery via vision language models. *ArXiv Preprint ArXiv:2509.01895*.

Esparza, M., Li, B., Ma, J., & Mostafavi, A. (2025). *International Journal of Disaster Risk Reduction AI meets natural hazard risk : A nationwide vulnerability assessment of data centers to natural hazards and power outages*. *126*(December 2024), 1–17. https://doi.org/10.1016/j.ijdrr.2025.105583

Farahmand, H., Liu, X., Dong, S., Mostafavi, A., & Gao, J. (2022). A Network Observability Framework for Sensor Placement in Flood Control Networks to Improve Flood Situational Awareness and Risk Management. *Reliability Engineering and System Safety*, *221*(January), 108366. https://doi.org/10.1016/j.ress.2022.108366

Farahmand, H., Xu, Y., & Mostafavi, A. (2023). A spatial – temporal graph deep learning model for urban flood nowcasting leveraging heterogeneous community features. *Scientific Reports*, 1–15. https://doi.org/10.1038/s41598-023-32548-x

Finney, Mark A. (2006). *An Overview of FlamMap Fire Modeling Capabilities*. 213–220.

Finney, Mark A, Grenfell, I. C., Mchugh, C. W., Seli, R. C., Trethewey, D., Stratton, R. D., & Brittain, S. (2011). *A Method for Ensemble Wildland Fire Simulation*. 153–167. https://doi.org/10.1007/s10666-010-9241-3

Finney, Mark Arnold. (1998). *FARSITE, Fire Area Simulator--model development and evaluation* (Issue 4). The Station.

Huot, F., Hu, R. L., Goyal, N., Sankar, T., Ihme, M., & Chen, Y. F. (2022). Next Day Wildfire Spread: A Machine Learning Dataset to Predict Wildfire Spreading From Remote-Sensing Data. *IEEE Transactions on Geoscience and Remote Sensing*, *60*, 1–13. https://doi.org/10.1109/TGRS.2022.3192974

Insights, E. (2025). *2025 LA County Wildfires*.

Insurance Institute for Business and Home Saftey. (2024). *The 2023 Lahaina Conflagration*. *September*.

Jain, P., Coogan, S. C. P., Subramanian, S. G., Crowley, M., Taylor, S., & Flannigan, M. D. (2020). A review of machine learning applications in wildfire science and management. *Environmental Reviews*, *28*(4), 478–505. https://doi.org/10.1139/er-2020-0019

Jiang, W., Wang, F., Su, G., Li, X., Wang, G., Zheng, X., Wang, T., & Meng, Q. (2022). Modeling Wildfire Spread with an Irregular Graph Network. *Fire*, *5*(6). https://doi.org/10.3390/fire5060185

Kadir, E. A., Kung, H. T., Almansour, A. A., Irie, H., Rosa, S. L., Sanim, S., & Fauzi, M. (2023). *Wildfire Hotspots Forecasting and Mapping for Environmental Monitoring Based on the Long Short-Term Memory Networks Deep Learning Algorithm*.

Knight, I., & Coleman, J. (1993). A Fire Perimeter Expansion Algorithm-Based on Huygens Wavelet Propagation. *International Journal of Wildland Fire*, *3*(2), 73–84. https://doi.org/10.1071/WF9930073

Luo, K., & Lian, I. Bin. (2024). Building a Vision Transformer-Based Damage Severity Classifier with Ground-Level Imagery of Homes Affected by California Wildfires. *Fire*, *7*(4). https://doi.org/10.3390/fire7040133

Ma, J., Li, B., Li, Q., Fan, C., & Mostafavi, A. (2025). *Attributed network embedding model for exposing COVID-19 spread trajectory archetypes*. 2693–2710. https://doi.org/10.1007/s41060-024-00627-5

Madaio, M., Chen, S.-T., Haimson, O. L., Zhang, W., Cheng, X., Hinds-Aldrich, M., Chau, D. H., & Dilkina, B. (2016). Firebird: Predicting fire risk and prioritizing fire inspections in Atlanta. *Proceedings of the 22nd ACM SIGKDD International Conference on Knowledge Discovery and Data Mining*, 185–194.

Mahmoud, H. (2024). Reimagining a pathway to reduce built-environment loss during wildfires. *Cell Reports Sustainability*, *1*(6), 100121. https://doi.org/10.1016/j.crsus.2024.100121

Mahmoud, H., & Chulahwat, A. (2018). Unraveling the Complexity of Wildland Urban Interface Fires. *Scientific Reports*, *8*(1), 1–12. https://doi.org/10.1038/s41598-018-27215-5

Mahmoud, H., & Chulahwat, A. (2020). Assessing wildland–urban interface fire risk. *Royal Society Open Science*, *7*(8), 201183. https://doi.org/10.1098/rsos.201183

Meldrum, J. R., Barth, C. M., Goolsby, J. B., Olson, S. K., Gosey, A. C., White, J., Brenkert-Smith, H., Champ, P. A., & Gomez, J. (2022). Parcel-Level Risk Affects Wildfire Outcomes: Insights from Pre-Fire Rapid Assessment Data for Homes Destroyed in 2020 East Troublesome Fire. *Fire*, *5*(1). https://doi.org/10.3390/fire5010024

Noumer, T. (2025). Predicting California Wildfire Damage to Structures Using Machine Learning : A Comparative Study of Random Forest and XGBoost Predicting California Wildfire Damage to Structures Using Machine Learning : A Comparative Study of Random Forest and XGBoost. *Journal of Physics: Conference Series*. https://doi.org/10.1088/1742-6596/3121/1/012029

Qiao, Y., Jiang, W., Su, G., Jiang, J., Li, X., & Wang, F. (2024). A transformer-based neural network for ignition location prediction from the final wildfire perimeter. *Environmental Modelling and Software*, *172*(August 2023), 105915. https://doi.org/10.1016/j.envsoft.2023.105915

Rahman, M. A., Hasan, S. T., & Kader, M. A. (2022). Computer Vision Based Industrial and Forest Fire Detection Using Support Vector Machine (SVM). *2022 International Conference on Innovations in Science, Engineering and Technology, ICISET 2022*, *February*, 233–238. https://doi.org/10.1109/ICISET54810.2022.9775775

Rösch, M., Nolde, M., Ullmann, T., & Riedlinger, T. (2024). Data-Driven Wildfire Spread Modeling of European Wildfires Using a Spatiotemporal Graph Neural Network. *Fire*, *7*(6). https://doi.org/10.3390/fire7060207

Rothermel, R. C. (1972). *A mathematical model for predicting fire spread in wildland fuels* (Vol. 115). Intermountain Forest & Range Experiment Station, Forest Service, US~….

Shirley, M. D. F., & Rushton, S. P. (2005). *The impacts of network topology on disease spread*. *2*, 287–299. https://doi.org/10.1016/j.ecocom.2005.04.005

Stratton, R. D. (2006). *Guidance on spatial wildland fire analysis: models, tools, and techniques*. United States Department of Agriculture, Forest Service, Rocky Mountain~….

Sullivan, A. L. (2009). Wildland surface fire spread modelling, 1990 - 2007. 3: Simulation and mathematical analogue models. *International Journal of Wildland Fire*, *18*(4), 387. https://doi.org/10.1071/wf06144

Taccaliti, F., Marzano, R., Bell, T. L., & Lingua, E. (2023). *Wildland – Urban Interface : Definition and Physical Fire Risk Mitigation Measures , a Systematic Review*.

Tehrany, M. S., Özener, H., Kalantar, B., Ueda, N., Habibi, M. R., Shabani, F., Saeidi, V., & Shabani, F. (2021). Application of an Ensemble Statistical Approach in Spatial Predictions of Bushfire Probability and Risk Mapping. *Journal of Sensors*, *2021*. https://doi.org/10.1155/2021/6638241

Tomy, A., Razzanelli, M., Di, F., Daniela, L., & Santina, C. Della. (2022). Estimating the state of epidemics spreading with graph neural networks. *Nonlinear Dynamics*, *109*(1), 249–263. https://doi.org/10.1007/s11071-021-07160-1

Veličković, P., Cucurull, G., Casanova, A., Romero, A., Lio, P., & Bengio, Y. (2017). Graph attention networks. *ArXiv Preprint ArXiv:1710.10903*.

Wu, W., Wang, J., Lu, K., Qi, W., Shan, F., & Luo, J. (2020). Providing service continuity in clouds under power outage. *IEEE Transactions on Services Computing*, *13*(5), 930–943. https://doi.org/10.1109/TSC.2017.2728795

Xiao, Y., Gupta, A., Esparza, M., Ho, Y.-H., Sebastian, A., Weas, H., Houck, R., & Mostafavi, A. (2025). Recov-vision: Linking street view imagery and vision-language models for post-disaster recovery. *ArXiv Preprint ArXiv:2509.20628*.

Xu, Z., Li, J., & Xu, L. (2024). *Wildfire Risk Prediction: A Review*. *2020*. http://arxiv.org/abs/2405.01607

Zhang, D., Roy, N., Wang, R., & Frost, J. D. (2025). *International Journal of Disaster Risk Reduction Predicting tornado-induced building damage : A comparative study of tree-based models and graph neural networks*. *123*(December 2024).

Zhao, Y., Gerard, S., & Ban, Y. (2024). *TS-SatFire: A Multi-Task Satellite Image Time-Series Dataset for Wildfire Detection and Prediction*. 1–16. http://arxiv.org/abs/2412.11555